# Deep neural networks for fine-grained surveillance of overdose mortality

Patrick J. Ward*[1,2], April M. Young[3,4], Svetla Slavova[1,2], Madison Liford[1], Lara Daniels[1], Ripley Lucas[3], Ramakanth Kavuluru[5,6]

1. Kentucky Injury Prevention and Research Center, University of Kentucky
2. Department of Biostatistics, University of Kentucky
3. Department of Epidemiology, University of Kentucky
4. Center on Drug and Alcohol Research, University of Kentucky
5. Department of Computer Science, University of Kentucky
6. Division of Biomedical Informatics, Department of Internal Medicine, University of Kentucky

*Corresponding Author

Email: pjward@email.unc.edu

## Abstract

Surveillance of drug overdose deaths relies on death certificates for identification of the substances that caused death. Drugs and drug classes can be identified through the International Classification of Diseases, 10th Revision (ICD-10) codes present on death certificates, however ICD-10 codes do not always provide high levels of specificity in drug identification. To achieve more fine-grained identification of substances on death certificate, the free-text cause of death section, completed by the medical certifier, must be analyzed. Current methods for analyzing free-text death certificates rely solely on look-up tables for identifying specific substances, which must be frequently updated and maintained. To improve identification of drugs on death certificates, a deep learning named-entity recognition model was developed, which achieved an F1-score of 99.13%. This model can identify new drug misspellings and novel substances that are not present on current surveillance look-up tables, enhancing the surveillance of drug overdose deaths.

## Introduction

Death certificates (DCs) are the primary data source for drug overdose (OD) mortality surveillance, providing information about the cause- and manner-of-death (1, 2). Medical certifiers write the cause-of-death in free-text fields on the DC (3, 4), which are then used to assign International Classification of Diseases, 10th Revision (5) (ICD-10) codes by the National Center for Health Statistics (NCHS) to DCs. ICD-10 codes are used at the national and local levels to calculate OD mortality statistics and to monitor trends in the drugs involved in ODs. Specifically, ICD-10 codes assigned as an underlying cause-of-death are used to identify drug OD deaths and those assigned as supplemental cause-of-death codes are used to identify the drug(s) involved in the OD (1, 3, 6). This process is central to the surveillance of drug OD mortality and is the primary way in which OD mortality information is reported to communities.

The ICD-10 classification process of OD deaths (called drug poisoning in ICD-10) has limitations, however, including timeliness of data (3, 7) and the lack of fine-grained granularity in the coding structure of ICD-10 for identification of specific drugs contributing to death (8). Some drugs have specific codes (e.g., T40.1 indicates heroin involvement, T40.3 indicates methadone involvement), but other drugs do not. For example, ICD-10 code T43.6 specifies poisoning by "psychostimulants with abuse potential (9)", which includes illicit substances such as methamphetamine and MDMA in addition to licit substances such as prescription amphetamine but excludes the psychostimulant cocaine, which has its own ICD-10 code (T40.5) (9). ICD-10 code T40.4 specifies "synthetic narcotics (2)", which includes fentanyl (prescribed or illicitly manufactured) and fentanyl analogs that have been a major cause of drug OD deaths in the United States (10-12), but also prescription drugs like tramadol.

This inexhaustive coding structure of ICD-10 has led epidemiologists to rely on text analyses to identify specific drugs involved in OD deaths when the codes are not indicative of a specific drug. In 2015, the Council of State and Territorial Epidemiologists (CSTE) released a tool to identify specific drug mentions from the free-text cause-of-death section of DCs (13). This tool consisted of a SAS program (13) that looped through individual words in the cause-of-death section and matched the words to a lookup table containing search terms (dug names, metabolites, and common misspellings) and a crosswalk to a referent drug. The tool produced a list of drugs found. In 2016, the NCHS developed a methodology for identifying specific drug involvement from the cause-of-death section that again relied on a lookup-table (3). This methodology additionally considered contextual information and had a more extensive table than the CSTE tool. While these tools presented improvements over using ICD-10 codes alone, they were limited by their lookup tables, requiring frequent updates with novel drugs. Additionally, maintaining every possible misspelling or metabolite for substances is a difficult, ongoing, and resource-intensive task.

Modern approaches within the field of natural language processing (NLP), including named entity recognition (NER) (14), could be applied in the area of drug OD surveillance to 1) expedite the process for identifying potential drug OD deaths before the ICD-10 coding of the DCs is completed; and 2) improve the identification of specific drug involvement by identifying drug search terms that are missing in the lookup tables.

NER involves tagging particular words as "named entities" (relevant to ODs, tagging substance names, misspellings, metabolites, and generics as "drug entities") and then training a machine learning (ML) algorithm to identify these entities from free-text. The trained algorithm can then predict what word(s) on new DC records are drug entities (15-17).

Modern NER techniques leverage deep learning (17-20) methods that involve artificial neural networks with several hidden layers. This allows for the model to automatically learn complex and robust features that are predictive of outcomes based on textual inputs (21). Manual feature engineering that was the hallmark of traditional NLP methods gave way to dense neural representations that have been shown to be more powerful in information extraction applications, including our current task. An advantage of this approach is that NER does not rely on a lookup table to determine which word(s) are indicative of a substance, rather, the ML algorithm through training learns which words (and surrounding contexts) are indicative of substances. NER has been used in clinical science to identify substances in free-text (15-18, 22), but has not been utilized for DCs to our knowledge.

The present study sought to advance the practice of drug OD mortality surveillance through the development of a modern NER tool for identifying substances on drug OD DCs. Integrating NER methods into the OD mortality surveillance pipeline will advance surveillance science (23) and enhance public health practice by improving the granularity and timeliness of OD mortality surveillance. This study demonstrates how adaptation of NER methodology enhances the current public health surveillance process of identifying specific drug involvement in ODs.

## Methods

*Data*

Data for this study was extracted from the Kentucky Death Certificate Database, Kentucky Office of Vital Statistics, Department of Public Health on March 9$^{th}$, 2020. All records coded with the consensus definition for drug OD death (1) for years 2014-2019 were identified and pulled from the database, resulting in a total of n=8,146 OD DC records used to develop the NER drug identification tool. Using R (24), the free-text data was parsed and reshaped so that every row was one individual element of the free-text from the cause-of-death section. Parsing did not remove any punctuation in the text as commas or hyphens may be parts of the name of a substance. Thus, punctuations were considered individual tokens. To demonstrate, a DC that said "HEROIN, FENTANYL, AND OXYCODONE OVERDOSE" was transformed into 7 rows (one for each token), corresponding to the elements "HEROIN", ",", "FENTANYL", ",", "AND", "OXYCODONE", "OVERDOSE". This process resulted in a dataset with 95,566 tokens. The text on the DCs is fully capitalized, and these tokens were used. The median number of tokens on a DC was 11 with a range of 1-75.

*Annotation*

To tag tokens as drug entities, the dataset was split in half. Two trained annotators labeled each split according to the beginning, inside, last, outside, unit (BILOU) tagging scheme (25). In the BILOU scheme, a single word that represents an entity is tagged as "U-*entity*" (unit). An entity that spans numerous tokens is tagged with "B-*entity*" (beginning), "I-*entity*" (inside), and "L-*entity*" (last), for the first token in the entity, any tokens between the first and last tokens in the entity, and the last token in the entity, respectively. Any words that do not represent an entity of interest are tagged with "O" (outside). BILOU was selected over other tagging schemes such as BIO as the annotators felt the BILOU scheme was more intuitive (and also more comprehensive enabling nested entities) for the annotation task than BIO. Table 1 shows how a record annotated with the BILOU scheme looks for drug entities. Kappa statistics were calculated for each split and for the entire dataset. Disagreements between annotators were identified and resolved at meetings of all four annotators to produce the final, annotated dataset.

*NER Drug Identification Tool and Modeling strategy*

Development of the NER tool utilized the Flair (26) library in Python. Flair is a flexible NLP library that provides a suite of word embeddings (27) in addition to a modeling framework for programming deep neural networks for NLP tasks, including NER. Word embeddings are semantic vector-space representations of words. In a typical embedding setting, words that are closely related have similar vector representations, which provides a more desirable quality for word-representation in NLP tasks compared to simple dictionary-key representations. Flair provides both previously developed embeddings (such as GloVe (27)) as well as Flair embeddings (26), which were developed using a character-level recurrent neural model that is contextualized by surrounding text, meaning an individual word can have multiple embeddings depending on the context. More recently, Flair embeddings were extended to “pooled Flair embeddings (28)”, to address issues with rare words. This is done by pooling (typically, averaging) embeddings from multiple occurrences of the same word across different instances and concatenating the pooled embedding to the contextual embedding of a particular instance under consideration.

In general, development of a model in the Flair framework follows the following steps: 1) labeled corpus creation, consisting of creating a train-test-validation split and pre-processing data to fit into Flair’s framework; 2) choosing word embeddings; 3) selecting a model; 4) training the chosen model using the selected embeddings. For this study, two models were developed using this framework, only differing in step 2. The n=8,146 DCs were split into training (n=6,108), validation (n=816), and test (n=1,222 DCs) sets, roughly a 75%-10%-15% split, respectively. This split was performed chronologically; DC records were sorted by date of death, and the first 75% were assigned to the training data, the next 10% were assigned to the validation data, and the final 15% were assigned to the test data. A chronological split was selected as it mimics the real-world use case for the model, where older data will predict new data. For word embeddings, a model with GloVe embeddings was developed as well as a model using forwards- and backwards-trained pooled Flair embeddings. Hypothetically, the contextual Flair embeddings should out-perform GloVe. Testing the pooled embeddings was of interest as DC text contains frequent misspellings as well as rare words such as metabolites of drugs and novel substances. The model development pipeline is displayed in Figure 1.

The modeling framework selected was a bidirectional long-short-term memory (BiLSTM) conditional random field (CRF) model. BiLSTM-CRF was selected as it achieved state-of-the-art performance in NER tasks previously (20). Flair contains a BiLSTM-CRF model which utilizes the PyTorch framework (29). For training, the model was set to run for 150 epochs. After each epoch, the model was tested on the validation set and the model’s accuracy was calculated. The highest-scoring model on the validation set was saved during training and deployed on the test data. Documented code for this process is available at https://github.com/pjward5656/DC_flair. Positive predictive value (PPV) (precision), sensitivity (recall), and F1-score (harmonic mean of PPV and sensitivity) (30) were calculated for each models’ performance on the test set for evaluation. High PPV indicates that there are few false positives; a PPV of 90% means that 9 out of every 10 entities the model identifies as drug entities are truly drug entities. In this example, the false positive rate is 10%. High sensitivity indicates that there are few false negatives; a sensitivity of 90% means that 9 out of every 10 tokens that are truly drug entities are correctly identified as drug entities. In this example, the false negative rate is 10%. F1-score provides one score to directly compare models. To demonstrate the advantage of the deep learning approach over other ML methods, a traditional CRF model that represents words as one-hot vectors (instead of as dense neural embeddings) was also developed to use as a baseline for comparison.

*Comparison to lookup table approach*

A widely used methodology for identification of specific drug involvement is based on literal text search for drug names, metabolites, and misspellings, cross-walked to a “referent drug” included in a lookup table of search terms. The methodology was implemented in a CSTE tool (3, 13) in 2015. The Kentucky Drug Overdose Fatality Surveillance System (31) has been updating the initial CSTE dictionary, and the current table includes more than 250 referent drugs (Appendix 1). To determine if the NER model provides an improvement over current methods, the lookup table approach was used to identify drug entities on the test data. Unlike the neural NER model developed, the lookup table approach has no false positives (any word identified from the lookup table is already confirmed to be a drug). The number of entities missed by the lookup table that were identified by the best BiLSTM-CRF was calculated to display how adding entities identified from the NER model could improve existing dictionaries.

## Results

### *Annotation*

The dataset had a total of n=95,566 tokens that required annotation. The kappa statistic between the first two annotators (first half of the dataset) was 0.996 and between the second two annotators (second half of the dataset) 0.973. Overall, the dataset had a kappa statistic of 0.983. The kappa values indicate the annotation task resulted in "perfect" inter-rater agreement level as per suggested rule of thumb (32), indicating a very high-quality dataset.

### *Modeling*

Table 2 displays the results from the ML approaches as well as the lookup table. While both deep learning models achieved high performance for this task, the model utilizing pooled Flair embeddings performed better, with both fewer false positives and false negatives, resulting in higher scores for Flair vs GloVe for PPV (99.16% vs 98.63%), sensitivity (99.10% vs 98.08%) and F1-score (99.13% vs 98.35%), although these differences are likely not statistically significant. Both deep learning models achieved higher scores than the baseline CRF model (F1-score=98.01%). Figure 2 displays the validation and training loss during training for the pooled Flair model.

The BiLSTM-CRF model using pooled Flair embeddings was able to recognize drug entities that the lookup table was not. In total, the deep learning approach identified 168 more entities than the lookup table. Appendix 2 displays the 130 unique drug search terms that the lookup table did not contain that were present in the test data. Many of these entities were terms that spanned multiple tokens, such as "DESIGNER OPIOIDS". While the lookup table had the term "OPIOIDS", the deep learning method identified the entity "DESIGNER OPIOIDS", which provides more information on the specific drug than the term "OPIOIDS" alone. Other missed entities included misspellings, such as "OXMORPHONE", "ALPRAZOLM", and "METHAMPHETTAMINE", which all have the correctly spelled terms in the lookup table but not these specific misspellings. Table 3 displays three DCs and shows the entities identified by each approach. By contrast, there were 13 instances where the opposite occurred, in which the lookup table identified a drug entity when the NER model did not.

## Discussion

### *Model Performance*

This study presents an accurate method for identifying drug entities on free-text drug OD DCs utilizing a modern NLP model. The F1-score of 99.13% for identifying drug entities achieved in the present study show promise for using deep learning methods in surveillance. The BiLSTM-CRF model leveraging pooled Flair embeddings identified more drug entities than the current lookup table for identifying drugs on free-text DCs, demonstrating that this method is an improvement over currently available tools. In total, the best deep learning approach identified 130 new unique drug entities compared to the lookup table. Despite the high performance, the model did make errors in some scenarios, including incorrectly classifying the word "METABOLITE" on several occasions. A more detailed error analysis is available in Appendix 3.

During training, the model achieved lower (better) loss on the validation data than the training data, as displayed in Figure 2. This is an unexpected result, as typically models will have higher scores on the in-sample training data than the out-of-sample validation set. Potential reasons for this include the relatively small sample size of the validation set (n=816) compared to the training set (n=6,108). Additionally, it is possible that the training data contained more difficult examples for the model to categorize than the validation data. A practical reason for this is how the data was split; the training data includes the oldest (chronologically) DC records. In Kentucky, public health initiatives have worked with medical certifiers to improve DC completion and accuracy for OD deaths, so DCs completed earlier in the epidemic may be of lower quality and thus more difficult for the model to learn than more recent DCs.

*Implications for public health practice*

While knowledge of deep learning methods and NER is not typically part of a public health practitioner's toolkit, epidemiologists working in OD surveillance possess knowledge of programming and text analytic tools as they have become required for identifying drugs that cause morbidity and mortality in communities (3). The high performance achieved by the developed model shows that deep learning should be leveraged moving forward in public health practice as a tool to solve challenges that appear in free-text data. Epidemiologists working at the state, local, and national levels should develop and implement these methods into surveillance pipelines to both improve public health practice and improve surveillance data quality. Neural NLP methods will become important as public health surveillance continues using more data sources that contain free-text, including electronic health records (33-36), emergency medical services run data (37-40), and syndromic surveillance (41-44), which are used for both drug ODs and other health conditions.

As many jurisdictions performing OD mortality surveillance will not have the expertise nor the computational power to develop and test NLP models, the developed model can advance surveillance efforts through improvement of the current surveillance pipeline. In analysis of Kentucky data alone, the deep learning model identified over 100 unique drug entities that a lookup table did not contain. Identified entities can be extracted from the model's results and added to the lookup tables of current tools, which can be disseminated to jurisdictions to improve identification of specific drugs. This will allow for the increased specificity of drug entities that the model provides without needing to run a complex model in situations where computational power or expertise is limited. With periodic runs on a fresh set of DCs, our model can thus surface new drug terms and improve operations in jurisdictions that do not necessarily have the resources to train and deploy neural models.

Integration of additional entities recognized by the model into existing tools in the OD mortality surveillance workflow will increase the specificity of drugs identified on OD DCs through fine-grained spotting, compared to both ICD-10 coding and other methods (3). Importantly, ongoing application of the NLP approach as part of the routine drug OD surveillance analysis will allow for the detection of novel substances as soon as they appear on a DC, before the lookup tables are updated with the new substance names. Addition of a new substance to a lookup table depends on circumstances within a rapidly changing drug market, which can be demanding. Since the developed algorithm can utilize context, it has the ability to recognize novel drug entities without constant modification of the underlying model. The model's ability to recognize novel entities will provide the opportunity for early warning signals for novel substances, and thus faster public health and public safety response. Additionally, the increased granularity will provide an overall improvement in surveillance data quality, which will lead to more accurate reporting of information to communities and stakeholders. The proposed neural NER methods expand our previous work on DC free-text (45) where we developed an NLP algorithm to capture drug OD death cases prior to ICD-10 coding. In combination, the two NLP algorithms can be added to the routine drug OD mortality surveillance tools to improve the early identification of drug OD deaths and emerging new drugs of concern, providing opportunities for timelier public health and safety response.

*Future research*

Future studies should apply the developed model to data from other jurisdictions and assess the model's performance to explore generalizability of the model. This generalizability would indicate that the model can be directly integrated into the surveillance work of other states and localities without the need for annotation and training. Additionally, future research should utilize a similar workflow for identifying substances on other surveillance data sources. While it is unlikely that the developed model, due to the nature of DC free-text, would achieve high scores on other free-text OD surveillance sources such as emergency medical services run data, future studies should explore the use of the Flair library and particularly pooled Flair embeddings for drug NER tasks. One limitation of the lookup table approach for identification of drugs involved in fatal ODs is the lack of context analysis. For example, if "history of heroin abuse" was mentioned on a drug OD DC, the lookup table approach as well as our currently proposed NLP algorithm would identify the OD death as heroin involved. Since NLP methods can be trained to recognize context, a future improvement of our NLP algorithm would be the filtering out of the drug entities in situations where the drugs were not mentioned as contributory to the OD death. This specific scenario did not occur frequently in Kentucky data used to develop the model, but will be an important improvement to avoid false positive cases for drug OD involvement of specific drugs. Future research should also explore simpler text matching methods, such as fuzzy matching, which may be more feasible for epidemiologists in resource-limited jurisdictions to implement.

*Strengths and limitations*

The developed method has several strengths. First, the dataset consisted of a large sample (n=8,146) of OD DCs spanning multiple years. The dataset was annotated by 4 trained annotators, ensuring accurate labels, verified by a high kappa statistic (0.983). Another strength of the method was the way in which the train-validation-test split was performed. By using earlier records for training and validation, and testing on later DCs, the model demonstrated that it can use older data to produce accurate predictions on new data, mimicking the real-world use-case of the model. The final strength of the model is the high F1-score (99.13%) achieved, displaying high accuracy for identifying drug entities on free-text DCs.

The study does have a few limitations. A primary limitation is that the developed NER model cannot be used as a stand-alone surveillance tool; rather, it should be implemented as an enhancement in current OD mortality surveillance work to improve drug identification on DCs. The lookup table-based methods contain crosswalks of drug search terms to their parent drug (for example, misspelling "CLONAZPAM" is cross-walked to parent drug "CLONAZEPAM"). Since the NER method, by design, recognizes novel entities, these entities are not present in dictionaries and their respective crosswalks, and therefore the NER model cannot be used as a surveillance tool alone. This limitation is addressed, however, by utilizing the model periodically on new data to recognize novel entities and adding these entities to existing lookup tables, so that novel entities can be identified by surveillance tools. Further, the inclusion of an entity in a lookup table guarantees that it will be recognized on every record it appears in, regardless of context—this addresses the rarely occurring scenario when a drug entity was not identified by the NER model due to it appearing in different context than the model learned.

An additional limitation is that the data used for training the model came from Kentucky DCs alone. The performance of the model on data from other jurisdictions should be evaluated. Additionally, the comparison of model performance to performance of the lookup table method is conditional on how up-to-date a given lookup table is; other jurisdictions may have more complete tables, so the performance improvement from the NER model may not be as high as in the present analysis. Another limitation is the complexity of the method and the need for computationally powerful hardware when applied to large datasets. Finally, the complexity of the model makes diagnosing errors difficult, which is an inherent limitation in most ML applications.

**Conclusion**

To our knowledge, this study is the first of its kind to use deep neural networks for drug NER on DCs. The highest performing model developed achieved an F1-score of 99.13%, indicating that the method is accurate at this task. The performance of the developed model shows that deep learning can be integrated into public health surveillance. Particularly for drug OD mortality, the method could improve surveillance data quality and timeliness, enabling public health practitioners to more quickly recognize novel substances and more accurately report data to communities. The developed method advances the science of public health surveillance by integrating NLP models not currently used in the field into surveillance and advances public health practice through enhancing data quality and timeliness of reporting. These surveillance improvements are key in monitoring the continuing drug OD epidemic and informing interventions to address this national crisis.

**Table 1: Example Annotated Death Certificate Free-text**

| Word | Tag |
|---|---|
| 7 | B-drug |
| - | I-drug |
| AMINOCLONAZEPAM | L-drug |
| AND | O |
| HEROIN | U-drug |
| OVERDOSE | O |

**Figure 1: Flair NLP model development and evaluation for overdose mortality surveillance**

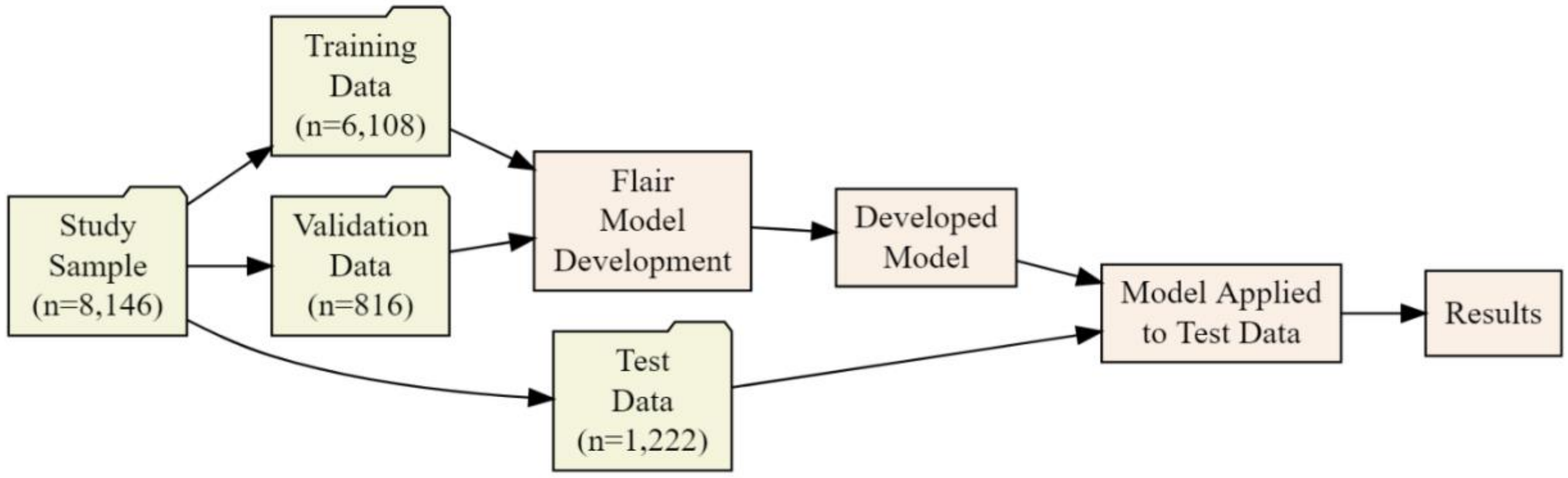

**Table 2: Performance of machine learning models on test set**

| Method Tested | True Positives | False positives | False negatives | PPV[1] | Sensitivity | F1-score |
|---|---|---|---|---|---|---|
| GloVe | 3168 | 44 | 62 | 98.63% | 98.08% | 98.35% |
| Pooled Flair | 3201 | 27 | 29 | 99.16% | 99.10% | 99.13% |
| CRF | 3135 | 32 | 95 | 98.99% | 97.05% | 98.01% |
| Lookup table | 3033 | 0 | 197 | 100.00% | 93.90% | 96.85% |

1. PPV: Positive predictive value

**Figure 2: Validation and Training Loss, Pooled Flair Model**

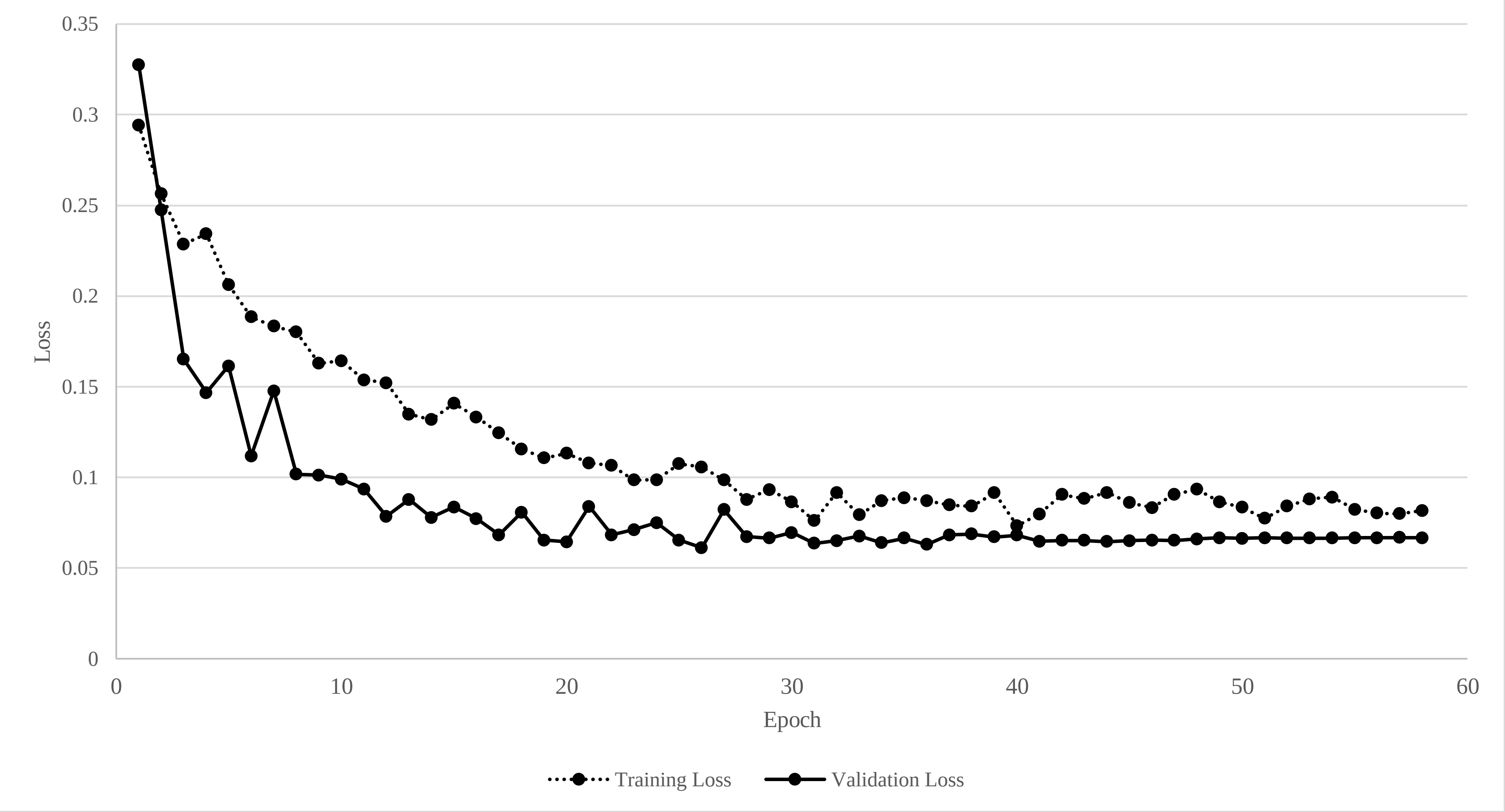

**Table 3: Example Classified Death Certificate Records**

| Death certificate tokens[1] | Recognized by the lookup table | Recognized by NER model[2] |
|---|---|---|
| ACUTE | | |
| INTOXICATION | | |
| BY | | |
| THE | | |
| COMBINED | | |
| EFFECTS | | |
| OF | | |
| FENTANYL* | ✓ | ✓ |
| , | | |
| ACETYLFENTANYL* | ✓ | ✓ |
| , | | |
| METHAMPHETTAMINE* | ✗ | ✓ |
| , | | |
| TRAMADOL* | ✓ | ✓ |
| , | | |
| AND | | |
| GABAPENTIN* | ✓ | ✓ |
| SUBSTANCE | | |
| ABUSE | | |
| MULTIPLE | | |
| DRUG | | |
| INTOXICATION | | |
| ( | | |
| METHAMPHETIMINE* | ✗ | ✓ |
| , | | |
| CLONAZAPAM* | ✗ | ✓ |
| , | | |
| OXYCODONE* | ✓ | ✓ |
| , | | |
| GABAPENTIN* | ✓ | ✓ |
| ) | | |
| MULTIPLE | | |
| DRUG | | |
| INTOXICATION | | |
| ACUTE | | |
| COMBINED | | |
| TOXIC | | |
| EFFECTS | | |
| OF | | |
| [ | | |
| METHAMPHETAMINE* | ✓ | ✓ |
| , | | |
| FENTANYL* | ✓ | ✓ |
| , | | |
| HEROIN* | ✓ | ✓ |
| , | | |
| GABAPENTIN* | ✓ | ✓ |
| , | | |
| & | | |
| PARAFLUOROBUTYRYLFENTANYL* | ✗ | ✓ |
| ] | | |
| SELF | | |
| ADMINISTRATION | | |
| OF | | |
| ILLICIT | | |
| DRUGS | | |

1. Tokens marked with an asterisk are drug entities; a horizontal line indicates the start of a new death certificate record.

**Appendix 1: Drug Overdose Fatality Surveillance System Drug Entities**

25B-NBOME
25C-NBOME
25D-NBOME
25H-NBOME
25I-NBOME
3METHYLFENTANYL
3-METHYLFENTANYL
3METHYLMORPHINE
3-METHYLMORPHINE
3-METHYLTHIOFENTANYL
4_ANPP
4ANPP
4-ANPP
4-METHOXYBUTYRYLFENTANYL
5F-AB-PINACA
5F-ADB
5F-ADB-PINACA
5F-AMB
5F-NNEI
5F-PB-22
5F-THJ
6-AM
6MAM
6-MAM
6-MOMOACETYLMORPHINE
6-MONACETYLMORPHINE
6-MONOACETYLMOPRHINE
6MONOACETYLMORPHINE
6-MONOACETYLMORPHINE
6-MONOACETYMORPHINE
6-MONOACETYTMORPHINE
6-MONOACEYTLYMORPHIN
6-MONOACEYTLYMORPHINE
6-MONOACEYTLYMORPHONE
6-MONOACTEYLMORPHINE
7-AMINOCLONAZEAPM
7AMINOCLONAZEPAM
7-AMINOCLONAZEPAM
AB-CHMINACA
AB-FUBINACA
AB-PINACA
ABSTRAL
JWH-073
JWH-122
JWH-210
JWH-250
LAMOTRIGINE
LAZANDA
LEVAMISOLE
LEVETIRACETAM
LEVOFLOXACIN
LIDOCAINE
LOPERAMIDE
LORAZEPAM
LORAZEPAN
LORCET
LORTAB
LYRICA
M-144
MAB-CHMINACA
MA-CHMINACA
MAM-2201
MARIJUANA
MATRIFEN
MAXIDONE
MDMA
MDMB-CHMICA
MDMB-FUBINACA
MDPV
MDPV
MEDICATION
MEPERIDINE
MEPHEDRONE
MEPHEDRONE
MEPROBAMATE
METAHDONE
META-METHYMETHOXYACETYL FENTANYL
METAXALONE
METFORMIN
METH
METHADONE
METHADOSE
METHAMHPETAMINE
METHAMPHETAMINE

ACETAMINOPHEN
ACETYL-ALPHA-METHYLFENTANYL
ACETYLFENTANYL
ACID
ACRYLFENTANYL
ACTIQ
ADB-FUBINACA
ADBICA
ADB-PINACA
ADDERALL
AH-7921
ALCOHOL
ALFENTANIL
ALPHA-METHYLFENTANYL
ALPHA-METHYLTHIOFENTANYL
ALPHA-PBP
ALPHA-PPP
ALPHA-PVP
ALPRAOZOLAM
ALPRAZOLAM
ALPRAZOLAN
ALPRAZOLEM
ALRPAZOLAM
AMB
AMBIEN
AMIDONE
AMINOCLONAZEPAM
AMIODARONE
AMITRIPTYLINE
AMLODIPINE
AMPHETAMINE
AMPHETAMINES
ANEXSIA
ANPP
ANTICONVULSANT
ANTICONVULSANTS
ANTIDEPRESSANT
ANTIDEPRESSANTS
ANTIPSYCHOTICS
A-OH-ALPRAOZOLAM
A-OH-ALPRAZOLAM
A-OH-ALPRAZOLAN
A-OH-ALPRAZOLEM
METHANOL
METHEDRONE
METHEDRONE
METHOCARBAMOL
METHORPHAN
METHOTREXATE
METHOXETAMINE
METHOXYACETYLFENTANYL
METHOXYBUTYRYLFENTANYL
METHYLENEDIOXYMETHAM
METHYLETHCATHINONE
METHYLETHCATHINONE
METHYLFENTANYL
METHYLONE
METHYLONE
METOCLOPRAMIDE
METONITAZENE
METOPROLOL
MIRTAZAPINE
MIRTAZEPINE
MITRAGYNINE
MN18
MO-CHMINACA
MOLLY
MOMOACETYLMORPHINE
MONACETYLMORPHINE
MONOACETYLMOPRHINE
MONOACETYLMORPHINE
MONOACETYMORPHINE
MONOACETYTMORPHINE
MONOACEYTLYMORPHIN
MONOACEYTLYMORPHINE
MONOACEYTLYMORPHONE
MONOACTEYLMORPHINE
MONOHYDROXYOXCARBAZEPINE
MONOXIDE
MOPRHINE
MORPHINE
MORPHONE
MT-45
NALOXONE
NALTREXONE
NARCOTIC

A-OH-ALRPAZOLAM
A-OH-APRAZOLAM
APP-FUBINACA
APRAZOLAM
A-PVP
ATENOLOL
BACLOFEN
BARBITURATE
BARBITURATES
BENADRYL
BENZODIAEPINE
BENZODIAEPINES
BENZODIAZEPINE
BENZODIAZEPINES
BENZOS
BENZOYLECGONINE
BENZOYLECGONINE
BENZTROPINE
BETA-HYDROXY-3-METHYLFENTANYL
BETA-HYDROXYFENTANYL
BETA-HYDROXYTHIOFENTANYL
BLEOMYCIN
BRORPHINE
BUPHEDRONE
BUPHEDRONE
BUPRENORFINE
BUPRENORPHINE
BUPROPION
BUPROPRION
BUSPIRONE
BUTALBITAL
BUTYLONE
BUTYLONE
BUTYRYLFENTANYL
CAFFEINE
CANNABINOIDS
CANNABIS
CARBAMAZEPINE
CARBAZEPINE
CARBON
CARFENTANIL
CARFENTANIL
NARCOTICS
NARCOTISM
N-DESMETHYL-TRAMADOL
NEURONTIN
NEUROSTIL
NICOTINE
NNEI
NORBUPRENORPHINE
NORCO
NORDIAZEPAM
NORDOXEPIN
NORFENTANYL
NORFLUOXETINE
NORPROPOXYPHENE
NORSERTRALINE
NORTRAMADOL
NORTRIPTYLINE
NORVENLAFAXINE
NUPENTIN
OCFENTANIL
O-DESMETHYL-TRAMADOL
OLANZAPINE
ONSOLIS
OPIATE
OPIATES
OPIOID
OPIOIDS
ORPHENADRINE
ORTHO-FLUORO FENTANYL
OXAZEAPM
OXAZEPAM
OXCARBAZEPINE
OXISET
OXYCODONE
OXYCODONE
OXYCONTIN
OXYCONTIN
OXYCOTIN
OXYGEN
OXYMOPHONE
OXYMORPHONE
OXYNORM

CARFENTANYL
CARISOPRODOL
CARISPRODOL
CHARCOAL
CHLORDIAZEPOXIDE
CHLORIDE
CHLOROPHENYLPIPERAZINE
CHLORPHENIRAMINE
CHLORPROMAZINE
CITALOPRAM
CLOMIPRAMINE
CLONAZEPAM
CLOZAPINE
COCAETHYLENE
COCAINE
CODEINE
CODEINE
COTININE
COUMADIN
CRYSTAL
CYANIDE
CYCLOBENZAPRINE
CYCLOPROPYLFENTANYL
DAMASON-P
DARVOCET
DARVOCET
DARVON
DEMEROL
DEPRESSANT
DESIPRAMINE
DESOMORPHINE
DEXTROMETHORPHAN
DEXTROPROPOXYPHENE
DIACETYLMORPHINE
DIAMORPHINE
DIAZDEPAM
DIAZEPAM
DICYCLOMINE
DIFLUOROETHANE
DIGOXIN
DIHYDROCODEINE
DIHYDROCODEINONE
DIHYDROMORPHINONE
OZAZEPAM
PALLADONE
PANACET
PARA-FLUOROBUTYRYLFENTANYL
PARA-FLUOROFENTANYL
PARA-FLUOROISOBUTYRYLFENTANYL
PARA-METHYMETHOXYACETYL FENTANYL
PAROXETINE
PB-22
PENRAL
PENTEDRONE
PENTEDRONE
PENTYLONE
PERCOCET
PERCODAN
PETNYLONE
PHARMACEUTICAL
PHARMACOLOGIC
PHENCYCLIDINE
PHENOBARBITAL
PHENTERMINE
PHENYTOIN
PIPERACILLIN
POLYDRUG
POLYPHARMACY
POLYSUBSTANCE
POTASSIUM
PREGABALIN
PRESCRIPTIONS
PROMETHAZINE
PROOXYPHENE
PROPAFENONE
PROPANE
PROPANOLOL
PROPOFOL
PROPOFOL
PROPOSYPHENE
PROPOXIPHENE
PROPOXITENE
PROPOXPHENE
PROPOXTYPHENE
PROPOXY
PROPOXYCODONE

DILAUDID
DILTIAZEM
DIPHENHYDRAMINE
DISKETS
DOLOPHINE
DOXEPIN
DOXYLAMINE
DULOXETINE
DURAGESIC
DURAGESIC
DUROGESIC
EAM-2201
ECSTASY
EDDP
ENDOCET
ENDODAN
EPHEDRINE
EPHEDRINE
ESCITALOPRAM
ETHABOL
ETHANOL
ETHYLENE
ETHYLMETHCATHINONE
ETHYLMETHOCATHINONE
ETHYLONE
ETHYLONE
ETIZOLAN
ETOH
ETOMIDATE
EXALGO
FAB-144
FANATREX
FDU-PB-22
FEENTANYL
FENATNYL
FENTANIL
FENTANLY
FENTANOL
FENTANY
FENTANYL
FENTANYL
FENTATYL
FENTAYNL
PROPOXYPHEN
PROPOXYPHENA
PROPOXYPHENE
PROPOXYPHERE
PROPOXYPHINE
PROPOXZPHENE
PROPRANOLOL
PROPXYPHENE
PSEUDOEPHEDRINE
PSYCHIATRIC
PX1
PX2
PX3
QUETIAPINE
REMIFENTANIL
ROXICET
ROXICODONE
ROXISET
SALICYLATE
SALICYLATES
SDB-006
SEROQUEL
SERTRALINE
SUBLIMAZE
SUBOXONE
SUBSTANCE
SUDAFED
SUFENTANIL
TEMAZEPAM
TETRAHYDROCANNABINOL
TETRAHYDROFURANFENTANYL
THC
THC-COOH
THIOFENTANYL
THJ
THJ-018
THJ-2201
TIZANADINE
TOBACCO
TOLUENE
TOPIRAMATE
TRAMADAL
TRAMADOL

FENTNAYL
FENTORA
FETANYL
FLEPHEDRONE
FLEPHEDRONE
FLUOROBUTYRYLFENTANYL
FLUOROISOBUTYRYLFENTANYL
FLUOXETINE
FLURAZEPAM
FURANYLFENTANYL
GABAPENTIN
GABAPIN
GABARONE
GABRION
GLYCOL
GRALISE
GUAIFENESIN
HALDID
HALOPERIDOL
HERION
HEROIN
HYCODAN
HYDROCADONE
HYDROCHLOROTHIAZIDE
HYDROCOCONE
HYDROCODINE
HYDROCODONE
HYDROCODONE
HYDROMORPHINE
HYDROMORPHONE
HYDROXYCHLOROQUINE
HYDROXYTHIOFENTANYL
HYDROXYZINE
HYSINGLA ER
IBUPROFEN
INSTANYL
ISOPROPANOL
ISOPROPYL
ISOTONITAZENE
JWH-015
JWH-018
JWH-019
TRAMADONE
TRAMDOL
TRAMEDOL
TRAMELL
TRAMIDOL
TRAMODOL
TRAZADONE
TRAZODONE
TREMEDEL
TREMEDOL
TUSSINEX
TYLENOL
TYLOX
U47700
U-47700
U-49900
U-51754
UR-144
VALERYL FENTANYL
VALIUM
VALPROIC
VENLAFAXINE
VERAPAMIL
VICIDAN
VICODIN
VICODIN
WARFARIN
XANAX
XANAX
XLR-11
XLR11
XLR12
XLR-12
XYLAZINE
ZOHYDRO ER
ZOLPIDEM
ZYDONE
NALMEFENE
QUININE
SUBLOCADE
SUBUTEX
NALBUPHINE

**Appendix 2: Drug entities present in test data not in Drug Overdose Fatality Surveillance System Table**

ALPRAZOLM
HYDROCONE
METHAMPHETAMINES
ACETALFENTANYL
METHAMPHETTAMINE
IMODIUM
OYCODONE
TETRAHYDROFURANFENTANYL
FENANTYL
NORIDIAZEPAM
ACETYFENTANYL
METHAMPETAMINE
INULIN
INSULIN
ANDACETYLFENTANY
HYDORCODONE
BENZO-DIAZEPINES
INDOMETHACIN
FENTANYLN
BENZODIAZAPENIES
4-NAPP
THC COOH
OPATES
CODIENE
4-ANNP
AMPHETAIMES
METHAMPHETIMINE
CLONAZAPAM
CANNABINOID
7-AMINOCIONAZEPAM
HYDROOXYZINECAUSING
ACETYFENTANTANYL
BUTYRYFENTANYL
TOPIRMATE
TCC-COOH
MORPINE
METHAMPH
EFFEXOR
PRISTIQ
FLEXERIL
OXMORPHONE
FLOUXETINE
NORFLUOXITINE
DEMOXEPAM
DESIGNER FENTANYLS
CANNABOIDS
4-AANP
ACETYLFENTAYL
DESIGNER OPOIDS
ALPRAZPOAM
BENZO
FENTANL
3,4 METHYLENEDIOXY-METHAMPHETAMINE
ALPROZOLAM
KETAMINE
METHOXYCETYFENTANYL
GAHAPENTIN
ARIPIPRAZOLE
DIPHENHYDRA
SERTALINE
VENAFLAXZINE
BENZOYLECGONINE.QUANT
NITROGLYCERIN
METHAMPHRTAMINE
CLONZEPAN
BENZODIAZIPINE
BENZODIAPINE
ZANAFLEX
METHOXYACETYLFENTANY
GGABAPENTIN
HYDROXYAINE
CONTININE
O-DESMETHYLVENLAFAXINE
MEXILETINE
NAPROXEN
FANTANYL
ACETYFENTANY
FLUXETINE
ACETLFENTANYL
ACETYFENTAN
ACETYLFENTANYL1
SODIUM NITRATE

$-ANPP
OXYCODE
METHAPHETAMINE
1,1-DIFLUOROETHANE
TCH-COOH
DESIGNER OPIOIDS
NIFEDIPINE
ACETYL FENTANYL
AMHPETAMINE
4-AMPP
PARAFLUROBUTYRYLFENTANYL
DESPROPIONYLFENTANYL
DESIGNER OPIATE
4-ANP
OXYCODEONE
METAMPHETAMINE
4ANNP
CRACK COCAINE
OXYCODONC
GABAPCNTIN
LOPERAMITE
IMMODIUM
OVERDOSE:FENTANYL
CALCIUM CHANNEL BLOCKER
BUSPRENORPHINE
WITHACETYLFENTANYL
COCAETHYENE
TRAZODON
HYDROX
METHANPHETAMINE
AMPHETATMINE
BENZODIAZOPINE
METHAMPHETEMINE
METHANPHETAMINES
ACETTYLFENTANYL
SODIUM NITRITE
4NAPP
ACEYLFENTANYL
METHAMPPHETAMINE
BUBRENORPHINE
METHAMPHATAMINE
METHAMPHETAMINR
DESIGNER DRUGS
GABAPENTIN_
DIIIIIAZEPAM
PARAFLUOROBUTYRYLFENTANYL
ACEYTLFENTANYL
PARA-FLUOROISOBUTYRYL FENTANYL

**Appendix 3: Error Analysis**

In total, 55 of the 17,450 total tokens in the test data were incorrectly classified by the deep learning model trained with pooled Flair embeddings. Most of these classification errors occurred under two circumstances: 1) the model identifying only part of an entity span as a drug but not all tokens in the entity and 2) the model incorrectly classifying the word "METABOLITE" as a substance. Situation 1) is common in any NER system where partial matches that are mostly correct are counted as errors. An example of this in our case is misclassification of entity span "CRACK COCAINE" by tagging the token "CRACK" as outside and the token "COCAINE" as an individual entity, when as per ground truth "CRACK" should be classified as the beginning of the entity and "COCAINE" as the end of the entity. This likely occurred for this entity specifically as "CRACK COCAINE" is not a commonly used terminology on DCs, considering that the model was able to correctly identify more common multi-token entities such as "DESIGNER OPIOIDS" or "DESIGNER FENTANYL" in other scenarios.

The second scenario involved the model classifying the word "METABOLITE" as a substance. An example of this occurring was on a DC that read "THC AND METABOLITE PRESENT". "METABOLITE" is not an entity of interest here, while "THC" is. A common context seen on DCs is lists of drugs, for example, "THC AND COCAINE PRESENT". Errors with the word "METABOLITE" likely occurred as the token appears in contexts where drugs typically appear, leading the model to misclassifying it as an entity.

Outside of these two described scenarios, other classification errors did not follow a specific pattern, besides instances where substances were misspelled extremely incorrectly (at times even incorrectly attached to other words, e.g. "HYDROOXYZINECAUSING"). These other errors are difficult to diagnose regarding reasons for why the model made mistakes. Errors like this did not occur frequently, however, as only a total of 55 tokens were misclassified in the test data.